%% file: aaai25.tex
\documentclass[letterpaper]{article} 
\usepackage{aaai25}  
\usepackage{times}  
\usepackage{helvet}  
\usepackage{courier}  
\usepackage[hyphens]{url}  
\usepackage{graphicx} 
\usepackage{natbib}  
\usepackage{caption} 
\usepackage{algorithm}
\usepackage{algorithmic}

\usepackage{newfloat}
\usepackage{listings}
\DeclareCaptionStyle{ruled}{labelfont=normalfont,labelsep=colon,strut=off} 
\floatstyle{ruled}
\newfloat{listing}{tb}{lst}{}
\floatname{listing}{Listing}
\usepackage{tabularx, booktabs}
\usepackage{multirow}
\usepackage{amsmath}
\usepackage{amssymb}
\usepackage{pifont}
\usepackage{enumitem}
\usepackage{xcolor,colortbl}

\definecolor{bgreen}{rgb}{0.0, 0.5, 0.0}
\definecolor{lightmint}{rgb}{0.92, 0.95, 0.98}

\title{AugRefer: Advancing 3D Visual Grounding via Cross-Modal Augmentation and Spatial Relation-based Referring}
\author{
    Xinyi Wang\textsuperscript{\rm 1},
    Na Zhao\textsuperscript{\rm 2}\thanks{Corresponding author.},
    Zhiyuan Han\textsuperscript{\rm 1},
    Dan Guo\textsuperscript{\rm 3},
    Xun Yang\textsuperscript{\rm 1}
}
\affiliations{
    \textsuperscript{\rm 1}University of Science and Technology of China\\
    \textsuperscript{\rm 2}Singapore University of Technology and Design\\
    \textsuperscript{\rm 3}Hefei University of Technology\\
    \{wxy1, aaronhan,xyang21\}@mail.ustc.edu.cn, na\_zhao@sutd.edu.sg, guodan@hfut.edu.cn}

\begin{document}
\maketitle

\begin{abstract}
3D visual grounding (3DVG), which aims to correlate a natural language description with the target object within a 3D scene, is a significant yet challenging task. 
Despite recent advancements in this domain, existing approaches commonly encounter a shortage: a limited amount and diversity of text-3D pairs available for training. Moreover, they fall short in effectively leveraging different contextual clues (\textit{e.g.}, rich spatial relations within the 3D visual space) for grounding. 
To address these limitations, we propose AugRefer, a novel approach for advancing 3D visual grounding. 
AugRefer introduces cross-modal augmentation designed to extensively generate diverse text-3D pairs by placing objects into 3D scenes and creating accurate and semantically rich descriptions using foundation models. Notably, the resulting pairs can be utilized by any existing 3DVG methods for enriching their training data. 
Additionally, AugRefer presents a language-spatial adaptive decoder that effectively adapts the potential referring objects based on the language description and various 3D spatial relations. 
Extensive experiments on three benchmark datasets clearly validate the effectiveness of AugRefer.
\end{abstract}

\section{Introduction}

\vspace{0.2cm}
3D visual grounding (3DVG) stands as an important and challenging task, aimed at locating objects within 3D scenes based on provided textual descriptions. As an advancement of 3D object detection \cite{zhao2020sess, sheng2022rethinking, zhao2022static, han2024dual, jiao2024unlocking}, it plays a critical perceptual role in various downstream applications, thus attracting increasing research attention.
The emergence of large language models (LLMs) adds further allure to this field, offering a pathway to connect LLMs with the physical 3D world seamlessly.

\begin{figure}[!t]
\centering
\includegraphics[width=0.45 \textwidth]{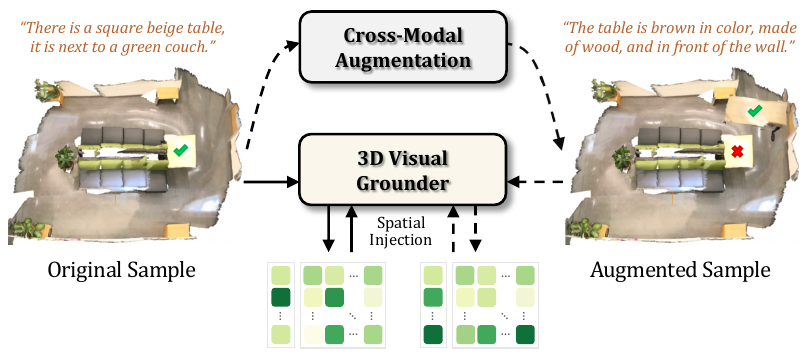} 
\caption{
A brief illustration of our proposed AugRefer: 1) \textbf{Cross-Modal Augmentation}: a brown wooden table is inserted into a living room scene, and generate its corresponding grounding description to increase data diversity. 2) \textbf{3D Visual Grounder}: we leverage spatial relation-based referring to grounding the target. }
\label{fig:teaser}
\end{figure}

Existing 3DVG methods commonly encounter a shortage of diverse training data pairs, consisting of the 3D scene with the referred object and the corresponding language description. 
This issue inherently arises from limitations on the 3D data side, where collecting and annotating 3D data is complex, costly, and time-consuming \cite{dai2017scannet, ding2023pla, distill-weakly-3dvg}. 
For example, the popular 3DVG dataset \cite{referit3d} only contains 1.5k scenes. 
Recent works (Hong et al. \citeyear{3d-llm}; Zhang et al. \citeyear{multi3drefer}) have used LLMs to enrich linguistic descriptions but have not addressed 3D data scarcity, while other studies (Ge et al. \citeyear{3dcopypaste}; Zhao et al. \citeyear{zhao2022synthetic}; Zhang et al. \citeyear{outdooraugmentation}) have explored 3D augmentation to introduce objects into existing scenes.
However, these single-modal augmentation techniques cannot be directly applied to cross-modal 3DVG due to two unique prerequisites inherent in augmenting text-3D pairs: 1) Ensuring \textbf{accurate correspondence} between 3D targets and linguistic descriptions, and 2) Providing \textbf{rich clues} necessary for locating the target, differentiating from other objects by incorporating both semantic and spatial information.

\begin{figure*}[t]
\centering
\includegraphics[width=0.86 \textwidth]{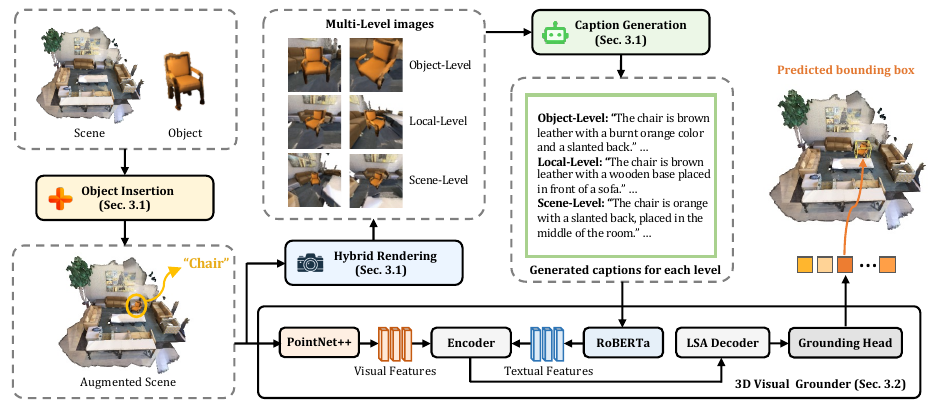} 
\caption{
\textbf{The framework overview of AugRefer.} It consists of two components: 1) \textbf{Cross-Modal Augmentation} with three steps: \textcircled{1} Object Insertion $\rightarrow$ \textcircled{2} Hybrid Rendering $\rightarrow$ \textcircled{3} Caption Generation; and 2) \textbf{3D Visual Grounder}, where our designed Language-Spatial Adaptive Decoder (LSAD) aims to enable more precise grounding by incorporating 3D spatial relations.}
\label{fig:augrefer_pipeline}
\end{figure*}

To tackle these limitations, we propose a novel approach for advancing 3D visual grounding, named AugRefer, as depicted in Fig.~\ref{fig:teaser}. In AugRefer, our initial step involves devising a cross-modal augmentation mechanism to enrich 3D scenes by injecting objects and furnishing them with diverse and precise descriptions.
This augmentation process involves three main steps: inserting objects into 3D scenes, rendering these scenes into 2D images, and using foundation models to generate detailed captions. Furthermore, we design a multi-granularity rendering strategy to capture intricate textures and tailored prompts to produce diverse captions for each level.
As a result, our cross-modal augmentation addresses the issue of data scarcity in 3DVG by significantly increasing the quantity and diversity of text-3D pairs.

In generated text-3D pairs, more complex situations arise. As shown in Fig.~\ref{fig:teaser}, if a scene already contains a table and our augmentation introduced an external table as the grounded target, the original one may become a distractor\footnote{Distractors refers to objects of the same category as the target.}, complicating the learning process. In such cases, it is essential to leverage spatial and other contextual information to make distinctions.
To date, many 3DVG methods either overlook the exploitation of valuable contextual clues \cite{butd-detr, eda, multiview-transformer}, such as rich spatial relations within the 3D visual space, or encounter challenges in effectively adapting the object features to different contextual clues \cite{3dvg-transformer,vil3dref, core-3dvg}.
In our AugRefer, we design a Language-Spatial Adaptive Decoder (LSAD) as the cross-modal decoder to facilitate more accurate grounding of the target object. The LSAD is engineered to adapt the features of potential target objects (\textit{i.e.}, the input to the decoder) to various contextual clues, including referring clues within the language description, object similarities in the 3D semantic space, and spatial relations within the 3D visual space.
Our LSAD explores two distinct types of spatial relations within the 3D visual space: \textit{global spatial relations} between objects and the entire scene as well as \textit{pairwise spatial relations} between objects as illustrated in Fig.~\ref{fig:lsad_framework} (a).  
Furthermore, we inject these spatial relations into the attention mechanism within the decoder in a novel manner. It's worth noting that our LSAD is compatible with any existing 3DVG framework employing a transformer-based architecture.

By integrating these two components, AugRefer achieves SOTA results on the ScanRefer and Sr3D datasets, showing the effectiveness of our proposed method. Furthermore, we demonstrate that our method can seamlessly integrate with existing 3DVG methods, such as BUTD-DETR and EDA, leading to consistent and significant improvements.

\section{Related Work}

\noindent{\textbf{3D Visual Grounding}}
focuses on locating the language-referred object in 3D point clouds, which is different from the 2D visual grounding~\cite{yang2021deconfounded,yang2022video}. 
Owing to advances in transformers \cite{transformer,pan2024finding}, the transformer-based methods have emerged as mainstream in 3DVG.
Most methods use attention mechanisms to fuse multi-modal features implicitly. For example, BUTD-DETR \cite{butd-detr} employs transformer-based encoder and decoder layers to fuse 3D visual features with features from other streams.
EDA \cite{eda} performs more fine-grained alignment between visual and textual features by decoupling the input text.
However, these standard attention modules lack the incorporation of spatial relationships.
To address this issue, 3DVG-Transformer \cite{3dvg-transformer} 
incorporates distance of proposals to capture pairwise spatial relations. 
CORE-3DVG \cite{core-3dvg} exploits the spatial features under the guidance of linguistic cues.
In this paper, we propose an effective spatial relation referring module for better global and pairwise perception.

\noindent{\textbf{3D Data Augmentation}}
aims to mitigate the challenge of data scarcity and significantly enhance performance.
Early efforts include but are not limited to, geometric transformations, noise injection, and generative methods.
Several recent indoor augmentation techniques also follow the practice of mixing samples in 3D outdoor detection tasks.
For instance, Mix3D \cite{mix3d} directly merges two point cloud scenes to achieve scene-level data augmentation, while DODA \cite{doda} creatively implements a cuboid-level merge between source and target point clouds, tailored for domain adaptation scenarios.
On the other hand, a few studies investigate the augmentation of 3D scenes with additional objects. 
For example, the outdoor 3D detection method Moca (Zhang et al. \citeyear{outdooraugmentation}) pastes ground-truth objects into both Bird's Eye View (BEV) and image features of training frames.
Likewise, 3D Copy-Paste \cite{3dcopypaste} inserts virtual objects into real indoor scenes.
In this work, we focus on implementing cross-modal augmentation between text descriptions and 3D scenes.

\section{Methodology}
\label{fig:method}

Consider a 3D indoor scene denoted by a 3D point cloud $P$ and a 
textual description $T$. Our goal is to predict the location  $B_t \in \mathbb{R}^6$ of the target 3D object based on the description $T$. 
To tackle the scarcity of 3D object-text pairs 
and incorporate object-to-object as well as scene-wide spatial context into object grounding, we propose a novel method, AugRefer, that performs cross-modal augmentation and spatial relation-based referring, as shown in Fig.~\ref{fig:augrefer_pipeline}.

\subsection{Cross-Modal Augmentation}
\label{sec:augmentation}
Our goal is to significantly diversify the limited 3D object-text training pairs. Our proposed cross-modal augmentation is a plug-and-play solution, easily integrable into existing models.
The process (as illustrated in Fig.~\ref{fig:augrefer_pipeline}) involves manipulating the 3D scenes in three key steps: 
\textbf{1) Insertion:} selecting suitable insertion positions for new objects and placing them in a plausible way that avoids interference with existing objects; 
\textbf{2) Rendering:} rendering the inserted objects in a multi-granularity way; 
and \textbf{3) Captioning:} taking the snapshots to generate diverse yet realistic descriptions. These captions are then refined to enhance their precision.

\subsubsection{Object Insertion.} 
\label{sec:insertion}
For a 3D scene and an external object, where the latter is randomly selected from other scenes, our first consideration is \textit{where}, \textit{what}, and \textit{how} to insert into the scene. 
To this end, we propose three main constraints for the Insertion operation: 1) \textit{the ground plane}, 2) \textit{the stander inserted object}, and 3) \textit{no collision with existing objects in the 3D scene}.
Algorithm 1 in the supplementary material outlines the plausible Insertion algorithm.

\begin{figure}[t]
\centering
\hspace*{-10pt}
\includegraphics[width=0.45 \textwidth]{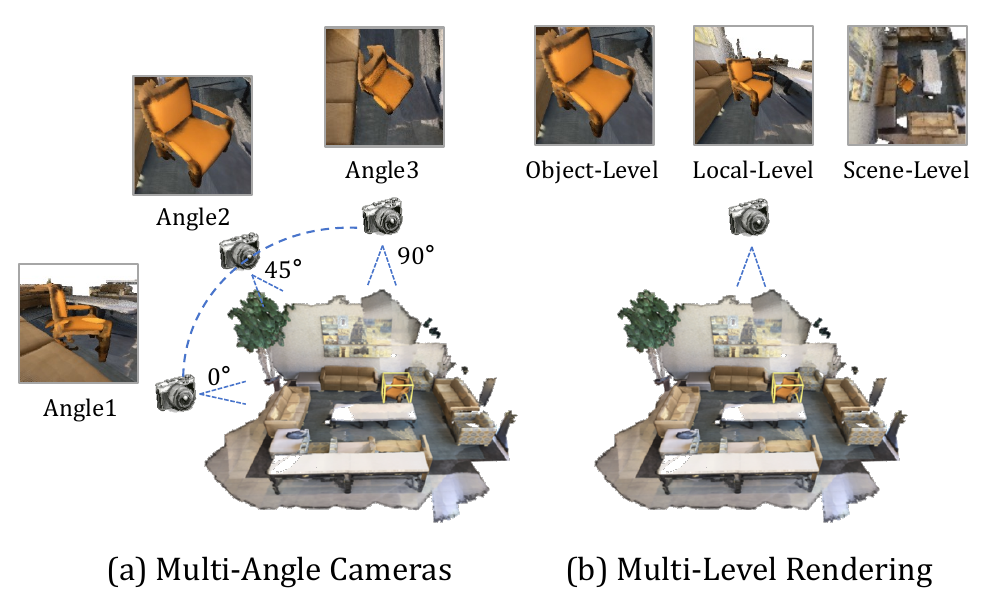} 
\caption{
\textbf{a) Multi-Angle Camera:} For each level of the scene, images are captured from multiple angles. \textbf{b) Multi-Level Rendering:} The scene is rendered at different levels.}
\label{fig:hybrid_rendering}
\end{figure}

In this step, we designate the floor as the primary area for introducing new elements, selecting the insertion plane with the smallest Z-axis value.
This entails imposing specific categorical constraints on the objects designated for insertion. Therefore, we focus on objects classified as a stander (\textit{e.g.}, table and chair) which naturally stands on the ground plane, rather than a hanger (\textit{e.g.}, window and curtain).
Subsequently, following (Zhao et al. \citeyear{zhao2022synthetic}), we simplify collision detection by converting the 3D scene into 2D floor images. Specifically, we employ an erosion technique in which the size of the object’s shape determines the kernel used to erode the ground plane, thereby identifying the collision-free area suitable for insertion.
If no suitable insertion area is found, we will resample another object and check the available space for insertion. This search process will continue until a viable insertion area is identified or the search limit is reached.
Finally, before the actual insertion, the object undergoes random jittering, flipping, and rotation along the Z-axis. 
The most frequently inserted stander objects during this process are chairs, cabinets, and tables.

\subsubsection{Hybrid Rendering.}
\label{sec:rendering}
A crucial step in our cross-modal augmentation involves linking augmented 3D scenes with appropriate descriptions. We achieve this by projecting the 3D scenes into 2D images and then generating rich and accurate descriptions through image captioning. To ensure the generation of high-quality descriptions, precise and visually detailed 2D images are essential. Therefore, we employ a hybrid rendering strategy that considers both multi-angle and multi-view aspects, as illustrated in Fig. \ref{fig:hybrid_rendering}.

Firstly, we develop a multi-angle camera placement strategy to address the occlusions present in the 3D point cloud. 
These occlusions arise from the cluttered nature of scenes and the inherent limitations of point cloud data collection, which often lead to incomplete object capture.
In our multi-angle camera placement strategy, we position cameras at 0, 45, and 90 degrees relative to the object's center and rotate them around the object to obtain multiple snapshots.
Secondly, we design a multi-level rendering strategy that encompasses object-, local-, and scene-level views to provide detailed attributes and spatial relationships of the inserted objects, as illustrated in Fig.\ref{fig:hybrid_rendering} (b).
Specifically, we render images at three levels of detail by centering the inserted object and adjusting the field of view: 1) \textit{object-level}: the object fills the frame, providing detailed insights into its categories and attributes. 2) \textit{local-level}: with a broader view showing the object's relationships with adjacent regions. 3) \textit{scene-level}: the view is expanded to include almost the entire scene for global contextual information.
Lastly, despite using multi-angle and multi-view rendering, issues like missing point clouds and obstructions can still arise and degrade image quality. To address this, we calculate CLIP \cite{clip} similarity scores between the images and their classes, selecting the top M images for the captioning phase.

\begin{figure}[t]
\centering
\hspace*{-10pt}
\includegraphics[width=0.48 \textwidth]{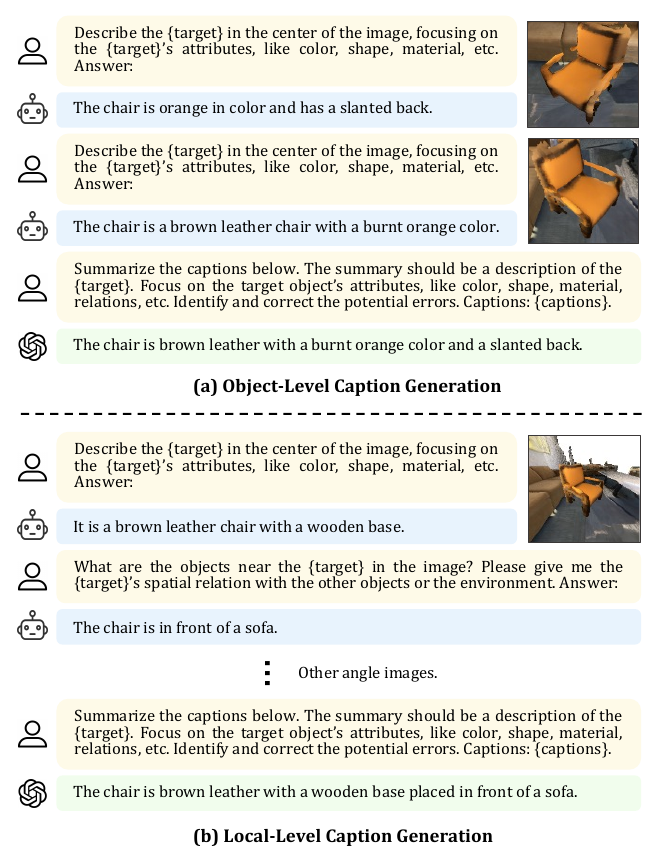} 
\caption{
\textbf{Multi-Level Caption Generation.} Conversation process with BLIP2 and ChatGPT for captioning various level rendering images. Both the Local-Level and Scene-Level captions utilize the same set of prompts. We describe the approach using the Local-Level as an example.
}
\label{fig:prompt}
\end{figure}

\subsubsection{Diverse Description Generation.}
\label{sec:captioning}
Building upon the success of 2D multi-modal pre-trained models \cite{blip2, gpt4}, we propose a strategy utilizing BLIP2 \cite{blip2} to generate accurate referential alignment. 
We meticulously craft various BLIP2 input prompts tailored for different levels of rendered images, as illustrated in Fig.~\ref{fig:prompt}. 
At the object-level, we instruct the model to provide detailed descriptions to capture finer visual characteristics. For the local and scene levels, we also require it to convey the spatial relationships between objects and their surroundings. Thus, we guide it to first identify the surrounding objects and then describe their interrelations.
In order to refine the captions at each level, we employ GPT-3.5 \cite{gpt3} to automatically identify and rectify potential inaccuracies prior to summarization. Additionally, we use GPT-3.5 to rephrase the captions, enhancing the diversity of descriptions and augmenting the textual modality.

\subsection{Overview of 3D Visual Grounder}
\label{sec:grounder}
Following existing approaches \cite{butd-detr, eda}, our 3D visual grounder consists of four basic modules: a feature extractor, a feature encoder, a cross-modal decoder, and a grounding head, as illustrated in Fig.~\ref{fig:augrefer_pipeline}.

\noindent\textbf{Feature Extractor.} 
We use a pre-trained PointNet++ \cite{pointnet++} to encode the input point cloud and extract visual features $\mathcal{V} \in \mathbb{R}^{N_p \times d}$. 
We adopt a pre-trained RoBERTa \cite{liu2019roberta} model to encode the textual input, generating language features $\mathcal{T} \in \mathbb{R}^{N_l \times d}$. Here, $N_p$ and $N_l$ denote the length of the visual and language tokens, respectively.
Following \cite{butd-detr}, we use box stream extracted from the GroupFree \cite{groupfree} detector to provide bounding box guidance for the visual features. 
We then utilize a learnable MLP layer to transform the $N_b$ detected bounding boxes into feature representation $\mathcal{B} \in \mathbb{R}^{N_b \times d}$.

\noindent\textbf{Feature Encoder.}
Within the encoder, visual and language features interact through the standard cross-attention layers, where they cross-attend to each other and to box proposal tokens using standard key-value attention in each layer.
Following the acquisition of cross-modal features $F_v \in \mathbb{R}^{N_p \times d}$ and $F_t \in \mathbb{R}^{N_l \times d}$, a linear layer is employed to select the top K visual features, denoted as $F_o \in \mathbb{R}^{K \times d}$, representing target object candidates.

\noindent\textbf{Cross-Modal Decoder.}
We design the decoder as a language-spatial adaptive decoder (LSAD), which further refines these candidates' visual features with the guidance of various contextual clues from different sources such as language, box stream, and visual information. 
We will discuss the details of LSAD in Sec.~\ref{sec:lsadecoder}.

\noindent\textbf{Grounding Head}.
The output object features of the decoder are fed into an MLP layer, which predicts the referential object bounding box. 
Following our baselines \cite{butd-detr, eda}, we respectively project visual and textual features into two linear layers, whose weights are denoted as $P_v \in \mathbb{R}^{K \times 64}$ and $P_t \in \mathbb{R}^{N_l \times 64}$, and compare the outputs with the ground-truths using soft-class prediction loss and semantic alignment loss.

\begin{figure}[!t]
\centering
\includegraphics[width=0.47 \textwidth]{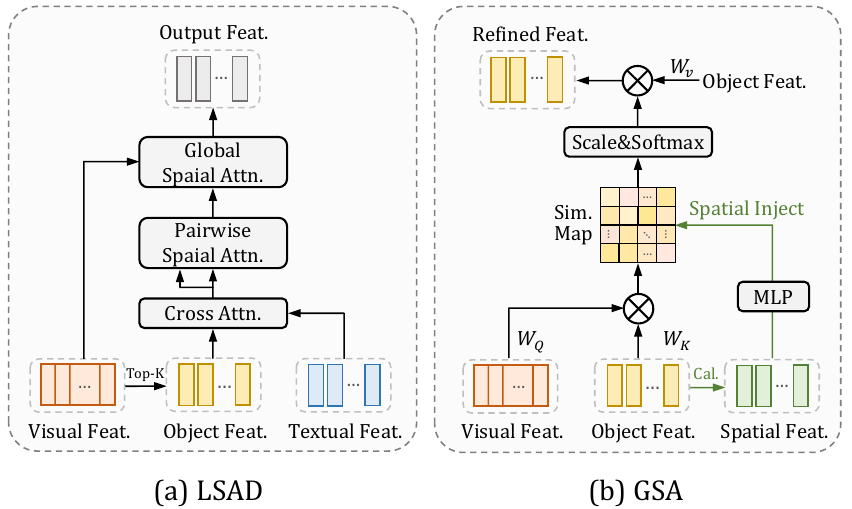}
\caption{
Illustrations of {a) Language-Spatial Adaptive Decoder (LSAD)} layer and {b) Global Spatial Attention (GSA)}.
}
\label{fig:lsad_framework}
\end{figure}

\subsection{Language-Spatial Adaptive Decoder}
\label{sec:lsadecoder}
The cross-modal decoder within our 3D visual grounder stands as the most critical module, tasked with modeling the contextual relationships between the target object and relevant objects that align with the language description, thereby facilitating the identification of the correct referred object.
Typically, the cross-modal decoder module, \textit{e.g.}, the decoder in \cite{butd-detr, eda}, utilizes cross-attention to capture the relationships between potential objects and language (or 3D object proposals) while applying self-attention to refine the features of candidate objects further. 
However, these conventional attention processes, which only model object-to-object relationships at a semantic level, cannot explicitly incorporate the spatial relationships between objects. Such spatial relationships are crucial for 3DVG \cite{3dvg-transformer, vil3dref, core-3dvg}, as the language description usually denotes objects based on their relative spatial positions within 3D scenes.

Furthermore, the inclusion of rich text-3D pairs generated by our cross-modal augmentation exacerbates the necessity of incorporating spatial relationships into the decoder. To address this, we introduce a Language-Spatial Adaptive Decoder (LSAD) designed to incorporate spatial relations from both global and pairwise perspectives.

The architecture of our LSAD layer is illustrated in Fig.~\ref{fig:lsad_framework} (a). In each decoder layer, we employ three distinct types of attention to refine the visual features of the objects. We initially perform cross-attention between object features and textual features, assigning weights to objects based on their relevance to the language. Subsequently, the objects engage in pairwise spatial attention to aggregate relative spatial relationships, followed by global spatial attention to gather global cues. After $N_D$ decoder layers, the final visual features are fed into the grounding head to predict the target.

\noindent\textbf{Global Spatial Attention.}
We design this module to achieve a better understanding of scene-wide spatial context, considering that global position descriptions also appear in the dataset, especially in our global-level augmented annotations, such as ``\textit{This speaker is brown \underline{in the corner}.}'' and ``\textit{This nightstand is \underline{in the middle}.}''.
Therefore, we introduce global spatial attention, which injects spatial relation information in the same manner as in pairwise spatial attention. 
However, the calculation of global spatial relationships and the attention targets differ.
Specifically, we calculate the normalized coordinates of the object center in the entire scene as global spatial features $R_g \in \mathbb{R}^{K \times 1 \times d_g}$:
\begin{equation}
r^g_i = [x_{\text{norm}}, y_{\text{norm}}, z_{\text{norm}}].
\end{equation}
Then we transform the global spatial relationship $R_g$ as $F_g$:
\begin{equation}
F_g = \text{MLP}(R_g).
\end{equation}
To improve the integration of global spatial relations, we refine object characteristic features $F_o$ by incorporating spatial features $F_g$ and the visual features of the entire scene point cloud $F_v$.
The process of global spatial attention (GSA) is outlined as follows:
\begin{equation}
\!Q \!=\! F_o W_Q, \ K = F_v W_K, \  V \!=\! F_v W_V, \  S_g = F_g W_S,
\end{equation}
\begin{equation}
\label{eqn:glb_attn}
\text{GSA}(Q, K, V) = \text{softmax}\left(\frac{Q K^T + S_g}{\sqrt{2 d_h}}\right) V,
\end{equation}
where $W_Q, W_K, W_V, W_S$ denote learnable linear layers.

\noindent\textbf{Pairwise Spatial Attention.}
In natural language descriptions, it is often necessary to distinguish the target from distractors by referring to one or more anchors and their pairwise spatial relationships, such as ``\textit{The chair \underline{next to} a brown couch}'' or ``\textit{There is a wooden cabinet \underline{between} a water cooler \underline{and} a trash can.}''. 
Therefore, we introduce Pairwise Spatial Attention (PSA), injecting spatial features into the same method as Global Spatial Attention. 
The calculation method of global relations and the features involved in the attention differ. 
Specifically, we calculate the distances and directions between objects to obtain the pairwise spatial relationships $R_p \in \mathbb{R}^{K \times K \times d_p}$, for $K$ objects, where $r^p_{ij}$ denotes the spatial relation between objects $O_i$ and $O_j$ (see the supplementary material for more details).

\tabcolsep=0.1cm
\begin{table*}
\begin{center}
\scalebox{0.88}{
\begin{tabular}{lcccccc}
    \toprule
    \multicolumn{1}{c}{\multirow{2}{*}{Method}} & \multicolumn{2}{c}{unique} & \multicolumn{2}{c}{multiple} & \multicolumn{2}{c}{overall} \\
    \cmidrule(r){2-3}
    \cmidrule(r){4-5}
    \cmidrule(r){6-7}
    & Acc@0.25 & Acc@0.5 & Acc@0.25 & Acc@0.5 & Acc@0.25 & Acc@0.5 \\
    \cmidrule(r){1-1}
    \cmidrule(r){2-3}
    \cmidrule(r){4-5}
    \cmidrule(r){6-7}
    ReferIt3D \footnotesize{\cite{referit3d}} & 53.75 & 37.47 & 21.03 & 12.83 & 26.44 & 16.90 \\
    ScanRefer \footnotesize{\cite{scanrefer}} & 67.64 & 46.19 & 32.06 & 21.26 & 38.97 &26.10 \\
    TGNN \footnotesize{\cite{tgNn}} & 68.61 & 56.80 & 29.84 & 23.18 & 37.37 & 29.70 \\
    SAT \footnotesize{\cite{sat}} & 73.21 & 50.83 & 37.64 & 25.16 & 44.54 & 30.14 \\
    FFL-3DOG \footnotesize{\cite{ffl-3dog}} & 78.80 & 67.94 & 35.19 & 25.70 & 41.33 & 34.01 \\
    InstanceRefer \footnotesize{\cite{instancerefer}} & 77.45  & 66.83 & 31.27 & 24.77 & 40.23 & 32.93 \\
    3DVG-Transformer \footnotesize{\cite{3dvg-transformer}} & 77.16 & 58.47 & 38.38 & 28.70 & 45.90 & 34.47 \\
    Multi-View Transformer \footnotesize{\cite{multiview-transformer}} & 77.67 & 66.45 & 31.92 & 25.26 & 40.80 & 33.26 \\
    3D-SPS \footnotesize{\cite{3d-sps}} & 81.63 & 64.77 & 39.48 & 29.61 & 47.65 & 36.43 \\
    Vil3DRef \footnotesize{\cite{vil3dref}} & 81.58 & 68.62 & 40.30 & 30.71 & 47.94 & 37.73 \\ 
    ViewRefer \footnotesize{\cite{viewrefer}} & 76.35 & 64.27 & 33.08 & 26.50 & 41.35 & 33.69 \\
    CORE-3DVG \footnotesize{\cite{core-3dvg}} & 84.99 & 67.09 & \underline{51.82} & \underline{39.76} & \underline{56.77} & 43.83 \\
    \bottomrule
    BUTD-DETR \footnotesize{\cite{butd-detr}} & 81.55 & 64.39 & 44.81 & 33.41 & 50.86 & 38.51 \\
    \rowcolor{lightmint} \textbf{+AugRefer}  &
    \textbf{85.21} \small\textcolor{bgreen}{+3.66} & 
    \textbf{68.99} \small\textcolor{bgreen}{+4.60} & 
    \textbf{47.73} \small\textcolor{bgreen}{+2.92} & 
    \textbf{37.16} \small\textcolor{bgreen}{+3.75} & 
    \textbf{53.91} \small\textcolor{bgreen}{+3.05} & 
    \textbf{42.41} \small\textcolor{bgreen}{+3.90} \\
    EDA \footnotesize{\cite{eda}} & 83.81	& 64.62	& 47.91 & 36.29 & 53.58 & 40.77 \\
    \rowcolor{lightmint} \textbf{+AugRefer}  &
    \underline{\textbf{86.21}} \small\textcolor{bgreen}{+2.40} & 
    \underline{\textbf{70.75}} \small\textcolor{bgreen}{+6.13} & 
    \textbf{49.96} \small\textcolor{bgreen}{+2.05} & 
    \textbf{39.06} \small\textcolor{bgreen}{+2.77} & 
    \textbf{55.68} \small\textcolor{bgreen}{+2.10} & 
    \underline{\textbf{44.03}} \small\textcolor{bgreen}{+3.26} \\
  \bottomrule
\end{tabular}}
\end{center}
\caption{Comparison with SOTA methods on \textit{ScanRefer}. We highlight the best performance with \underline{underlining}.}
\label{tab:benchmark_scanrefer}
\end{table*}

\section{Experiments}
\label{sec:experiments}

\subsection{Dataset and Experimental Setting}
\noindent\textbf{Datasets.}
We use three 3DVG datasets: ScanRefer (Chen et al. \citeyear{scanrefer}), Nr3D \cite{referit3d}, and Sr3D \cite{referit3d} to evaluate our method. 
Note that Sr3D and Nr3D provide ground-truth objects, and some methods simplify the grounding task to a matching problem of selecting the ground-truth box that best matches the description.
Following \cite{core-3dvg}, we use detected objects as input with the raw point cloud, instead of ground truths.

\noindent\textbf{Evaluation Metrics.} 
We evaluate performance using the Acc@k (k=0.25 or 0.5) metric.
Acc@k means the accuracy where the best-matched proposal has an intersection over union with the ground truth greater than the threshold k.

\noindent\textbf{Baselines.}
We choose two 3DVG models, specifically BUTD-DETR \cite{butd-detr} and EDA \cite{eda} as baselines\footnote{Since the SOTA method CORE-3DVG is not open source, we use the top-performing models EDA and BUTD-DETR in 3DVG. Note that our method could be compatible with CORE-3DVG.
}.  
In our experiments, we integrate our cross-modal augmentation and hierarchy spatial decoder into the baselines, respectively, to verify the effectiveness of our method.

\noindent\textbf{Implementation Details.}
Our experiments are conducted on four NVIDIA A100 80G GPUs, utilizing PyTorch and the AdamW optimizer. We adjust the batch size to 12 or 48 and augment training with 22.5k generated pairs for each dataset. 
On average, the generated description contains 13.7 words.
The visual encoder's learning rate is set to 2e-3 for ScanRefer, while other layers are set to 2e-4 across 150 epochs. In contrast, SR3D and NR3D have learning rates of 1e-3 and 1e-4, respectively; NR3D undergoes 200 epochs of training, whereas SR3D requires only 100 epochs due to its simpler, template-generated descriptions.

\begin{table}[t]
\begin{center}
\scalebox{0.88}{
\begin{tabular}{lcc}
    \toprule
    \multicolumn{1}{c}{\multirow{2}{*}{Method}} & \multicolumn{1}{c}{Nr3D} & \multicolumn{1}{c}{Sr3D} \\
    \cmidrule(r){2-2}
    \cmidrule(r){3-3}
    & Acc@0.25 & Acc@0.25 \\
    \cmidrule(r){1-1}
    \cmidrule(r){2-2}
    \cmidrule(r){3-3}
    ReferIt3D \footnotesize{\cite{referit3d}} \dag & 24.00 & 27.70 \\
    InstanceRefer \footnotesize{\cite{instancerefer}} \dag & 29.90 & 31.50 \\
    LanguageRefer \footnotesize{\cite{languagerefer}} \dag & 28.60 & 39.50 \\
    SAT \footnotesize{\cite{languagerefer}} \dag & 31.70 & 35.40 \\
    CORE-3DVG \footnotesize{\cite{core-3dvg}} & \underline{49.57} & 54.30 \\
    \bottomrule
    BUTD-DETR \footnotesize{\cite{butd-detr}}  & 38.60 & 53.64 \\
    \rowcolor{lightmint} \textbf{+AugRefer}  &
    \textbf{48.41} \small\textcolor{bgreen}{+9.81} & 
    \underline{\textbf{60.22}} \small\textcolor{bgreen}{+6.58} \\
    EDA \footnotesize{\cite{eda}} & 42.08 & 51.39 \\
    \rowcolor{lightmint} \textbf{+AugRefer}  &
    \textbf{46.49} \small\textcolor{bgreen}{+4.41} & \textbf{57.95} \small\textcolor{bgreen}{+6.56} \\
  \bottomrule
\end{tabular}
}\end{center}
\caption{Comparison with SOTA methods on \textit{Nr3D} and \textit{Sr3D}. $\dag$ Evaluation results are quoted from \cite{core-3dvg}. We highlight the best performance with \underline{underlining}.} 
\label{tab:benchmark_referit3d_all}
\end{table}

\tabcolsep=0.09cm
\begin{table}[!t]
\begin{center}
\scalebox{0.9}{
\begin{tabular}{l||cc||cc|cc|cc}
   \hline\hline
    \multicolumn{1}{c||}{\multirow{2}{*}{\-}} &
    \multicolumn{2}{c||}{Method} &
    \multicolumn{2}{c|}{unique} & \multicolumn{2}{c|}{multiple} & \multicolumn{2}{c}{overall} \\
    & CA & LSAD & @0.25 & @0.5 & @0.25 & @0.5 & @0.25 & @0.5 \\
    \hline
    (a) & \- & \-  & 81.55 & 64.39 & 44.81 & 33.41 & 50.86 & 38.51 \\
    (b) & \ding{51} & \- & 85.13 & 66.62 & 46.40 & 34.96 & 52.79 & 40.18 \\
    (c) & \- & \ding{51} & 84.52 & 67.92 & 46.85 & 36.03 & 53.07 & 41.29 \\
    (c) & \ding{51} & \ding{51} & \textbf{85.21} & \textbf{68.99} & \textbf{47.73} & \textbf{37.16} & \textbf{53.91} & \textbf{42.41} \\
     \hline\hline
\end{tabular}
}\end{center}
\caption{The ablation study of our AugRefer. CA stands for cross-modal augmentation; LSAD for language-spatial adaptive decoder. BUTD-DETR is used as baseline (row a).}
\label{tab:two_strategy}
\end{table}

\subsection{Overall Comparison}
In Tab. \ref{tab:benchmark_scanrefer} and Tab. \ref{tab:benchmark_referit3d_all}, we report the comparison of performance between the baselines and our AugRefer, as well as the comparison between ours and the reported SOTA results, across three 3DVG datasets, \textit{i.e.} ScanRefer, Nr3D, and Sr3D. 
This leads us to the following insights:
\begin{itemize}[leftmargin=0.4cm]
    \item Our method exhibit significant improvements across various metrics when integrated into the baselines, BUTD-DETR and EDA. In particular, when compared to BUTD-DETR, our performance improvements \textit{w.r.t.} the overall Acc@0.25 metric are 3.05\%, 9.81\%, and 6.58\%, respectively on ScanRefer, Nr3D, and Sr3D. Furthermore, by integrating EDA, the open-source model with the best performance, AugRefer significantly increases accuracy by 2.10\%, 4.41\%, and 6.56\% across the three datasets.
    \item Benefiting from our AugRefer, EDA has achieved the SOTA level in Nr3D and Sr3D datasets. In Scanrefer, we have brought EDA closer to the level of SOTA. It's noteworthy that our AugRefer is also compatible with the SOTA model CORE-3DVG \cite{core-3dvg}, offering the potential for further performance enhancements.  
\end{itemize}

\tabcolsep=0.3cm
\begin{table}[tb]
\begin{center}
\scalebox{0.94}{
\begin{tabular}{l|ccc|cc}
    \hline\hline
    \multicolumn{1}{c|}{\multirow{2}{*}{\-}} &
    \multicolumn{3}{c|}{Level} &
    \multicolumn{2}{c}{overall} \\
    & object & local & scene & @0.25 & @0.5 \\
    \hline
    (a) & \- & \- & \- & 50.86 & 38.51 \\
    (b) & \ding{51} & \- & \- & 48.34 & 36.67 \\
    (c) & \- & \ding{51} & \- & 49.32 & 37.06 \\
    (d) & \- & \- & \ding{51} & 50.38 & 38.74 \\
    (e) & \- & \ding{51} & \ding{51} & 51.04 & 38.70 \\
    (f) & \ding{51} & \ding{51} & \ding{51} & \textbf{52.79} & \textbf{40.18} \\
   \hline\hline
\end{tabular}
}\end{center}
\caption{The ablation study of augmentation levels.
}
\label{tab:aug_level}
\end{table}

\tabcolsep=0.3cm
\begin{table}[tb]
\centering
{
\begin{tabular}{l|c|cc}
      \hline\hline
    \multicolumn{1}{c|}{\multirow{2}{*}{\-}} &
    \multicolumn{1}{c|}{Quantity per} &
    \multicolumn{2}{c}{overall} \\
    & scene and level & @0.25 & @0.5 \\
    \hline
    (a) & 1  & 51.49 & 39.15 \\
    (b) & 3  & \textbf{52.79} & \textbf{40.18} \\
    (c) & 5  & 51.37 & 38.38 \\
    \hline\hline
\end{tabular}
}
\caption{The ablation study on augmentation quantities.}
\label{tab:quantity_comparison}
\end{table}

\begin{figure*}[t]
\centering
\includegraphics[width=0.88 \textwidth]{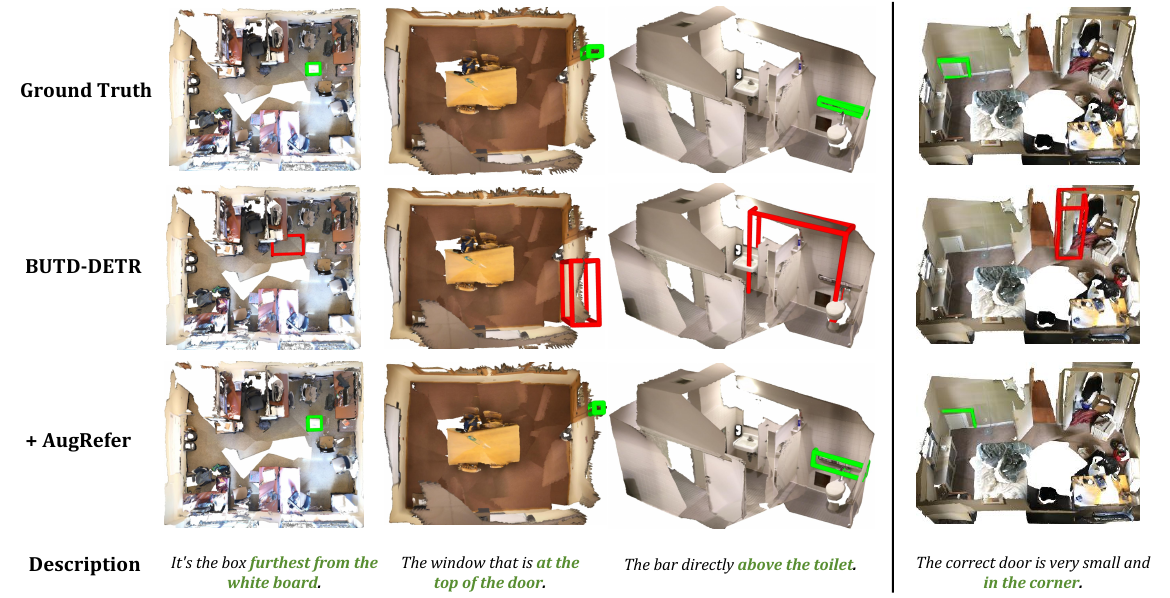}
\caption{
Qualitative results with ScanRefer descriptions: (a) ``\textit{It's the box furthest from the whiteboard}.''  (b) ``\textit{The brown door is in the corner of the room}.''
}
\label{fig:qualitive_results}
\end{figure*}

\begin{table}[t]
\centering
\tabcolsep=0.3cm
{
\begin{tabular}{l|c|cc}
      \hline\hline
    \multicolumn{1}{c|}{\multirow{2}{*}{\-}} &
    \multicolumn{1}{c|}{\multirow{2}{*}{Method}} &
    \multicolumn{2}{c}{overall} \\
    & \- & @0.25 & @0.5 \\
    \hline
    (a) & pair-glb-lang & 51.18 & 39.27  \\
    (b) & pair-lang-glb & 52.93 & 41.02  \\
    (c) & lang-pair-glb & \textbf{53.91} & \textbf{42.41} \\
    \hline\hline
\end{tabular}
}
\caption{The ablation study on language priors and visual priors in the LSAD module. 
}
\label{tab:spatial_framework}
\end{table}

\subsection{In-depth Studies}
\textbf{Two strategies synergize to make AugRefer effective.}
The results in Tab. \ref{tab:two_strategy} show that our cross-model augmentation significantly boosts performance in the simple splits, \textit{i.e.}, `\textit{unique}',  owing to the inclusion of augmented 3D scenes and a broader array of augmented objects, which enhance the model's ability to perceive object classes.
In contrast, spatial injection yields greater improvements in the more challenging splits, \textit{i.e.}, `\textit{multiple}'. This stems from the learning of the injected spatial relations, thereby strengthening the model's capability to differentiate distractors according to spatial relations.
Together, these two strategies synergize to form the rational AugRefer.
Overall, these findings highlight the complementary benefits of our proposed approaches in improving 3D visual grounding performance.

\noindent\textbf{Multi-level enhances accuracy.}
We apply multi-level rendering and caption generation strategies in our cross-modal augmentation. To investigate the effect of different levels in the cross-augmentation, we gradually introduce distinct levels of augmented samples into the baseline BUTD-DETR and report the results in Tab. \ref{tab:aug_level}.
It is interesting to note that while each level of augmentation alone does not improve overall performance compared to the baseline without augmentation (row a), combining all three levels of granularity in augmentation allows AugRefer to achieve improvements of 1.93\% and 1.67\% in overall metrics. This highlights the importance of our multi-level design, which captures detailed attributes and spatial relationships of newly inserted objects, providing more precise descriptions.
In addition, we also conduct experiments to investigate the relationship between the quantity of generated pairs and performance improvements. During the training phase of BUTD-DETR baseline, we randomly add $n$ pairs ($n = 1, 3, 5$) from each scene and level to augment the training dataset. 
We observed that increasing the number of generated pairs can lead to performance degradation, likely due to noise introduced in the generated pairs. To maintain a balance between generated and original pairs, we set $n$ to 3.

\noindent\textbf{Our current LSAD design outperforms the alternatives.}
In Tab. \ref{tab:spatial_framework}, we investigate the effects of order in applying three types of attention on ScanRefer dataset with cross-modal augmented training data. 
Rows (a-c) incorporate the LSAD module with varying orders of the three attentions. The results in Tab. \ref{tab:spatial_framework} show that modeling complex spatial relationships is more effective when language priors are applied (cross-attended) first. 
Additionally, we compare our LSAD module with an alternative design that also models spatial relationships to highlight the advantages of our LSAD. Specifically, we replace the LSAD decoder in our baseline BUTD-DETR with the corresponding structure from Vil3DRef \cite{vil3dref}, denoted as `Vil. Decoder' in Tab. \ref{tab:spatial_comparison}. The comparison results show that our LSAD module, which integrates both global and pairwise relationships, achieves significantly better performance.

\tabcolsep=0.3cm
\begin{table}[t]
\centering
{
\begin{tabular}{l|c|cc}
      \hline\hline
    \multicolumn{1}{c|}{\multirow{2}{*}{\-}} &
    \multicolumn{1}{c|}{\multirow{2}{*}{Method}} &
    \multicolumn{2}{c}{overall} \\
    & \- & @0.25 & @0.5 \\
    \hline
    (a) & BUTD-DETR  & 50.86 & 38.51 \\
    (b) & \- +Vil. Decoder \- & 50.18 & 38.65 \\
    (c) & +LSAD & \textbf{53.91} & \textbf{42.41} \\
    \hline\hline
\end{tabular}
}
\caption{Comparison of the spatial relation module.}
\label{tab:spatial_comparison}
\end{table}

\noindent{\textbf{Qualitative Analysis}}: 
Fig. \ref{fig:qualitive_results} illustrates the qualitative comparison between AugRefer and the baseline BUTD-DETR using two samples from ScanRefer dataset. 
The visualization results demonstrate that our AugRefer outperforms the baseline, particularly in grounding challenging objects such as the small box.
Furthermore, AugRefer exhibits notable improvements in spatial modeling, particularly in the pairwise (first three columns) and global (last column) contexts.

\section{Conclusion}

Our work alleviates the shortage of both amount and diversity in the text-3D grounding dataset and the inefficiencies of exploring contextual clues by introducing AugRefer.
We enrich 3D scenes with additional objects and generate detailed descriptions in three distinct granularities using foundation models.
Furthermore, we integrate contextual clues into the model, enabling a thorough comprehension of these relationships.
This paper pioneers the use of cross-modal augmentation techniques, substantially advancing the field of 3D visual grounding and providing viable solutions wherever in research and practical fields. In the future, we aim to extend our idea to tackle more complex reasoning tasks~\cite{yang2024robust,Luo2025TraME}.

\section*{Acknowledgments}
This research work was supported by the National Natural Science Foundation of China (NSFC) under Grant U22A2094, and also supported by the Agency for Science, Technology and Research (A*STAR) under its MTC Programmatic Funds (Grant No. M23L7b0021). We also acknowledge the support of the advanced computing resources provided by the Supercomputing Center of the USTC, and the support of GPU cluster built by MCC Lab of Information Science and Technology Institution, USTC.

\bibliography{aaai25}

\clearpage
\newpage
\renewcommand\thesection{\Alph{section}}
\renewcommand\thesubsection{\thesection.\arabic{subsection}}
\setcounter{section}{0}
\input{aaai25-Supp}

\end{document}

%% file: aaai25-Supp.tex



\section{Implementation Details}
\label{sec:implementation}
\noindent\textbf{Object Insertion Algorithm.} 
Section \ref{sec:augmentation} of the main paper discusses our strategy, further detailed in Algorithm \ref{alg:insertion}.
For each indoor scene in \cite{dai2017scannet}, we plausibly place objects in 3D scenes by selecting the ground plane, enforcing categorical constraints, and ensuring collision-free areas through 2D conversion and erosion techniques, followed by random transformation before final insertion.

\begin{algorithm}[h]
\caption{Augmented Object Insertion}
\label{alg:insertion}
\textbf{Input}: A scene $S_i$ from 3D indoor dataset $S$, its floor map $f$ \\
\textbf{Output}: Augmented scene $\hat s$, inserted object $\hat{o}$ and its location $\hat{b}$ (center and size)
\begin{algorithmic}[1] 
\STATE \textbf{Initialize:}: $T \gets 1000$, $\hat{s} \gets None$, $\hat{o} \gets None$
\FOR{$j \in \{1, 2, \ldots, T\}$}{
    \STATE randomly choose another scene $S_j$
    \STATE randomly select a stander object $o$ from scene $S_j$
    \STATE Calculate the size of $o$ and the eroded floor map $\hat{f}$ 
    \IF{$\hat{f} == \emptyset$ and $j < T$}
        \STATE \textbf{continue} 
    \ELSE
        \STATE randomly jitter, flip, and rotate $o$ along the Z-axis
        \STATE insert at a random location $\hat{b}$
        \STATE $\hat{s} \gets \hat{s_i} \oplus o_j $, $\hat{o} \gets {o_j} $ 
        \STATE break
    \ENDIF}
\ENDFOR
\STATE Return $\hat{s}$, $\hat{o}$, $\hat{b}$
\end{algorithmic}
\end{algorithm}

\noindent\textbf{Feature Encoder.}
We use hyperparameters consistent with those in the baseline BUTD-DETR \cite{butd-detr} and EDA \cite{eda}. The encoder receives three input streams: visual features $\mathcal{V} \in \mathbb{R}^{N_p \times d}$, textual features $\mathcal{T} \in \mathbb{R}^{N_l \times d}$, and box features $\mathcal{B} \in \mathbb{R}^{N_b \times d}$. Here, $N_p=1024$, $N_b=133$, $d=288$, and $N_l$ denotes the maximum length of language tokens in a batch. After $N_E=3$ encoder layers, we select the top K target object candidates along with their features $F_o \in \mathbb{R}^{K \times d}$, where $K=256$.

\noindent\textbf{Cross-Modal Decoder.}
After calculating spatial relationships, we obtain $R_p \in \mathbb{R_p}^{K \times K \times 5}$ and $R_g \in \mathbb{R}^{K \times 1 \times 3}$, which subsequently are projected to $S_p \in \mathbb{R}^{K \times K \times d}$ and $S_g \in \mathbb{R}^{K \times 1 \times d}$. After $N_D=6$ decoder layers, we get the output features $F_v' \in \mathbb{R}^{K \times d}$, which incorporate spatial information.

\noindent\textbf{Pairwise Spatial Attention.}
For each pair of objects $O_i$ and $O_j$, 
we calculate their Euclidean distance, as well as the horizontal and vertical sine and cosine values of the lines connecting the object centers, and finally concatenate them together into the spatial pairwise spatial realation vector $r^p_{ij}$, as described in \cite{vil3dref}.
We then map them to pairwise spatial features, denoted as $S_p$ = MLP ($R_p$).
The calculation method for Pairwise Spatial Attention (PSA) follows a similar approach to Global Spatial Attention (GSA) in the main paper. Specifically, Q, K, and V are all derived from the object features $F_o$, with $S_p$ replacing $S_g$, while the rest of the process remains unchanged.

\noindent\textbf{Augmentation and Model Training.}
We generate a set of augmented 3D-text grounding pairs using our cross-modal augmentation technique. We perform ten distinct object insertions at various levels for each scene and produce precise linguistic descriptions accordingly.
For joint detection prompts in the baseline BUTD-DETR \cite{butd-detr} and EDA \cite{eda}, we incorporate the category label of the inserted object into the prompts. 
During the 3DVG training phase, we randomly select three samples from each scene and level to expand the training dataset, resulting in approximately 22.5k additional augmented training pairs.
Therefore, GPU memory usage remains constant, while the training time increases proportionally, with extra 10 hours required on 4 Nvidia A100 GPUs. Inference time remains unchanged.

\section{Additional Results}
\label{sec:result}
\vspace{0.1cm}
We provide a detailed performance comparison of our method against the baselines BUTD-DETR and EDA on Nr3D and Sr3D \cite{referit3d} datasets. The results, presented in Tab. \ref{tab:result_nr3d} and \ref{tab:result_sr3d}, demonstrate that our method outperforms the baselines across all splits, underscoring its effectiveness and broad applicability.

\tabcolsep=0.15cm
\begin{table}[!ht]
{
\begin{tabular}{c|ccccc}
    \hline\hline
    \multicolumn{1}{c|}{\multirow{2}{*}{Method}} &
    \multicolumn{1}{c}{\multirow{2}{*}{Easy}} & 
    \multicolumn{1}{c}{\multirow{2}{*}{Hard}} & 
    \multicolumn{1}{c}{View-} & \multicolumn{1}{c}{View-} & \multicolumn{1}{c}{\multirow{2}{*}{Overall}} \\
    & & & -dep. & indep. & \\
    \hline
    BUTD-DETR & 45.25 & 31.89 & 33.76 & 40.52 & 38.60 \\
    \textbf{+AugRefer}
    & \textbf{54.80}
    & \textbf{41.97} 
    & \textbf{40.31} 
    & \textbf{51.63}  
    & \textbf{48.41}  \\
    & \textbf{\small\textcolor{bgreen}{+9.55}} 
    & \textbf{\small\textcolor{bgreen}{+10.08}}
    & \textbf{\small\textcolor{bgreen}{+6.55}} 
    & \textbf{\small\textcolor{bgreen}{+11.11}}
    & \textbf{\small\textcolor{bgreen}{+9.81}} \\
    \hline
    EDA & 46.69 & 37.52 & 35.25 & 45.52 & 42.08 \\
    \textbf{+AugRefer}
    & \textbf{52.41} 
    & \textbf{40.64}  
    & \textbf{38.80} 
    & \textbf{50.37} 
    & \textbf{46.49}  \\
    & \textbf{\small\textcolor{bgreen}{+5.72}} 
    & \textbf{\small\textcolor{bgreen}{+3.12}}
    & \textbf{\small\textcolor{bgreen}{+3.55}} 
    & \textbf{\small\textcolor{bgreen}{+4.85}} 
    & \textbf{\small\textcolor{bgreen}{+4.41}} \\
  \hline\hline
\end{tabular}
}
\centering
\caption{Detailed comparison with SOTA methods on \textit{Nr3D}.}
\label{tab:result_nr3d}
\end{table}

\vspace{-0.1in}
\tabcolsep=0.15cm
\begin{table}[!h]
{
\begin{tabular}{c|ccccc}
    \hline\hline
    \multicolumn{1}{c|}{\multirow{2}{*}{Method}} &
    \multicolumn{1}{c}{\multirow{2}{*}{Easy}} & 
    \multicolumn{1}{c}{\multirow{2}{*}{Hard}} & 
    \multicolumn{1}{c}{View-} & \multicolumn{1}{c}{View-} & \multicolumn{1}{c}{\multirow{2}{*}{Overall}} \\
    & & & -dep. & indep. & \\
    \hline
    BUTD-DETR \dag & 56.05 & 47.98 & 44.28 & 54.06 & 53.64 \\
    \textbf{+AugRefer}
    & \textbf{63.51}
    & \textbf{52.51} 
    & \textbf{51.73} 
    & \textbf{60.60}  
    & \textbf{60.22}  \\
    & \textbf{\small\textcolor{bgreen}{+7.46}}
    & \textbf{\small\textcolor{bgreen}{+4.53}}
    & \textbf{\small\textcolor{bgreen}{+7.45}} 
    & \textbf{\small\textcolor{bgreen}{+6.54}}
    & \textbf{\small\textcolor{bgreen}{+6.58}} \\
    \hline
    EDA \dag & 53.96 & 45.35 & 45.89 & 51.63 & 51.39 \\
    \textbf{+AugRefer}
    & \textbf{60.33} 
    & \textbf{52.38} 
    & \textbf{54.79} 
    & \textbf{58.09} 
    & \textbf{57.95} \\ 
    & \textbf{\small\textcolor{bgreen}{+6.37}} 
    & \textbf{\small\textcolor{bgreen}{+7.03}}
    & \textbf{\small\textcolor{bgreen}{+8.90}} 
    & \textbf{\small\textcolor{bgreen}{+6.46}} 
    & \textbf{\small\textcolor{bgreen}{+6.56}} \\
  \hline\hline
\end{tabular}
}
\centering
\caption{Detailed comparison with SOTA methods on \textit{Sr3D}.}
\label{tab:result_sr3d}
\end{table}

\section{Qualitative Analysis}
\label{sec:vis}
\begin{figure*}[ht]
\centering
\includegraphics[width=0.9 \textwidth , height=0.45 \textwidth]
{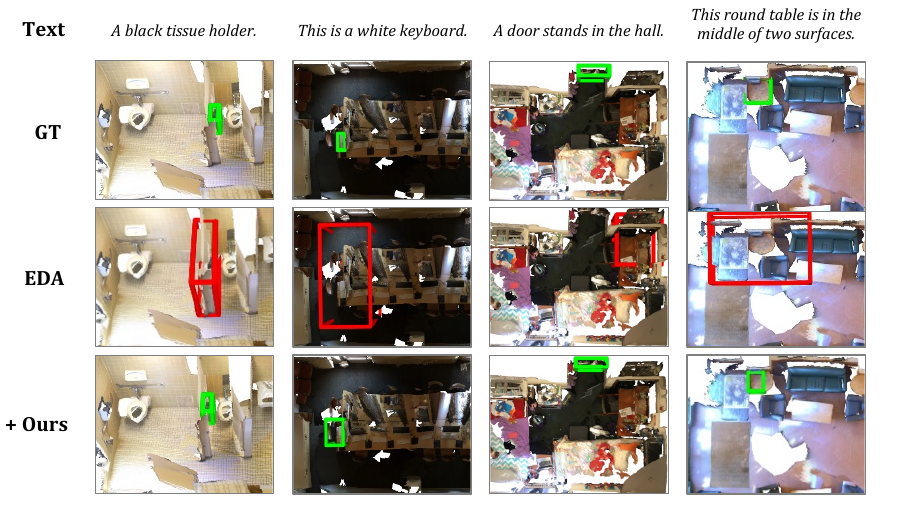} 
\caption{
\normalfont
\textbf{
Qualitative comparison on samples from ScanRefer dataset.} 
}
\label{fig:vis_scanrefer}
\end{figure*}

\begin{figure*}[ht]
\centering
\includegraphics[width=0.9 \textwidth , height=0.45 \textwidth]
{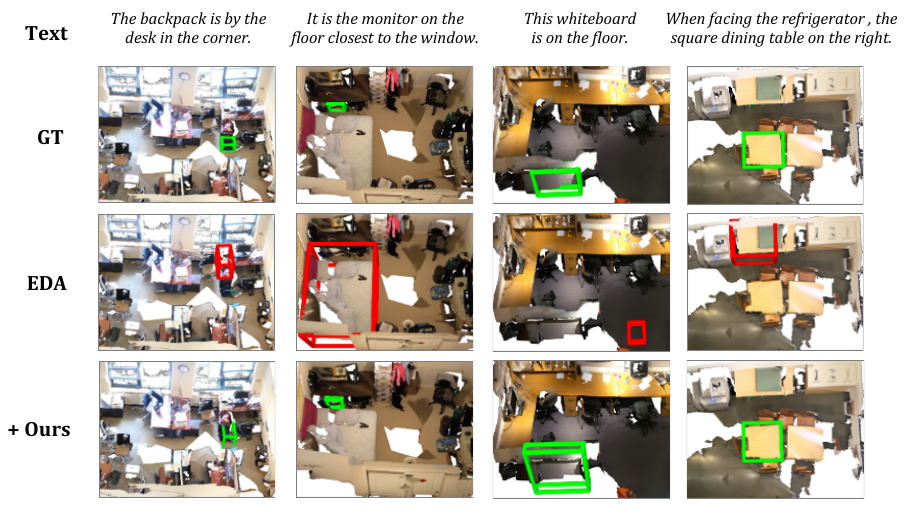} 
\caption{
\normalfont
\textbf{
Qualitative comparison on samples from Nr3D dataset.} 
}
\label{fig:vis_nr3d}
\end{figure*}

\begin{figure*}[ht]
\centering
\includegraphics[width=0.9 \textwidth , height=0.45 \textwidth]
{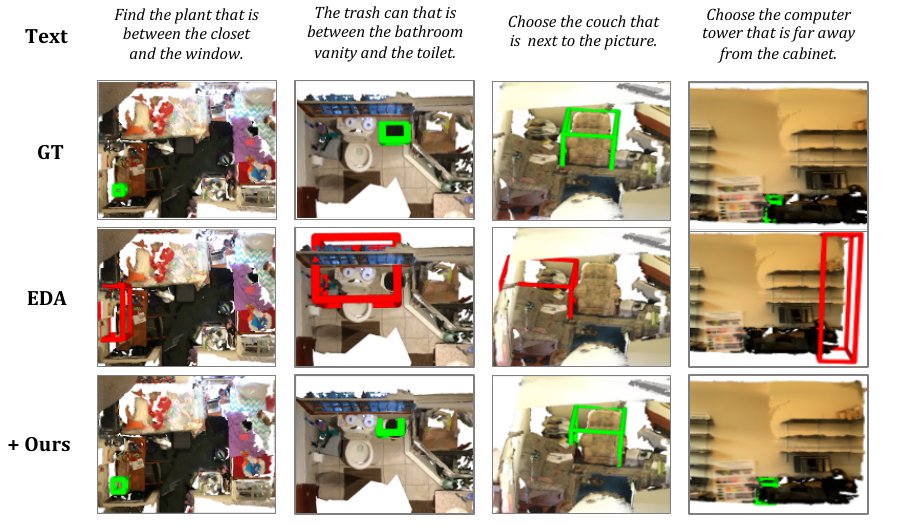} 
\caption{
\normalfont
\textbf{
Qualitative comparison on samples from Sr3D dataset.} 
}
\label{fig:vis_sr3d}
\end{figure*}

\vspace{0.1cm}
Qualitative results on three datasets ScanRefer \cite{scanrefer}, Nr3d and Sr3d \cite{referit3d} for 3D vision grounding task are shown in Fig. \ref{fig:vis_scanrefer}, \ref{fig:vis_nr3d} and \ref{fig:vis_sr3d}.
Compared to the top-performing baseline EDA, our method offers more precise perception and positioning for the given description, whether in object category, appearance, or spatial relationship.

%% file: aaai25.bbl
\begin{thebibliography}{43}
\providecommand{\natexlab}[1]{#1}

\bibitem[{Achiam et~al.(2023)Achiam, Adler, Agarwal, Ahmad, Akkaya, Aleman, Almeida, Altenschmidt, Altman, Anadkat et~al.}]{gpt4}
Achiam, J.; Adler, S.; Agarwal, S.; Ahmad, L.; Akkaya, I.; Aleman, F.~L.; Almeida, D.; Altenschmidt, J.; Altman, S.; Anadkat, S.; et~al. 2023.
\newblock Gpt-4 technical report.
\newblock \emph{arXiv preprint arXiv:2303.08774}.

\bibitem[{Achlioptas et~al.(2020)Achlioptas, Abdelreheem, Xia, Elhoseiny, and Guibas}]{referit3d}
Achlioptas, P.; Abdelreheem, A.; Xia, F.; Elhoseiny, M.; and Guibas, L. 2020.
\newblock Referit3d: Neural listeners for fine-grained 3d object identification in real-world scenes.
\newblock In \emph{Computer Vision--ECCV 2020: 16th European Conference, Glasgow, UK, August 23--28, 2020, Proceedings, Part I 16}, 422--440. Springer.

\bibitem[{Brown et~al.(2020)Brown, Mann, Ryder, Subbiah, Kaplan, Dhariwal, Neelakantan, Shyam, Sastry, Askell et~al.}]{gpt3}
Brown, T.; Mann, B.; Ryder, N.; Subbiah, M.; Kaplan, J.~D.; Dhariwal, P.; Neelakantan, A.; Shyam, P.; Sastry, G.; Askell, A.; et~al. 2020.
\newblock Language models are few-shot learners.
\newblock \emph{Advances in neural information processing systems}.

\bibitem[{Chen, Chang, and Nie{\ss}ner(2020)}]{scanrefer}
Chen, D.~Z.; Chang, A.~X.; and Nie{\ss}ner, M. 2020.
\newblock Scanrefer: 3d object localization in rgb-d scans using natural language.
\newblock In \emph{European conference on computer vision}, 202--221. Springer.

\bibitem[{Chen et~al.(2022)Chen, Guhur, Tapaswi, Schmid, and Laptev}]{vil3dref}
Chen, S.; Guhur, P.-L.; Tapaswi, M.; Schmid, C.; and Laptev, I. 2022.
\newblock Language conditioned spatial relation reasoning for 3d object grounding.
\newblock \emph{Advances in neural information processing systems}, 35: 20522--20535.

\bibitem[{Dai et~al.(2017)Dai, Chang, Savva, Halber, Funkhouser, and Nie{\ss}ner}]{dai2017scannet}
Dai, A.; Chang, A.~X.; Savva, M.; Halber, M.; Funkhouser, T.; and Nie{\ss}ner, M. 2017.
\newblock Scannet: Richly-annotated 3d reconstructions of indoor scenes.
\newblock In \emph{Proceedings of the IEEE conference on computer vision and pattern recognition}, 5828--5839.

\bibitem[{Ding et~al.(2022)Ding, Yang, Jiang, and Qi}]{doda}
Ding, R.; Yang, J.; Jiang, L.; and Qi, X. 2022.
\newblock Doda: Data-oriented sim-to-real domain adaptation for 3d semantic segmentation.
\newblock In \emph{European Conference on Computer Vision}, 284--303. Springer.

\bibitem[{Ding et~al.(2023)Ding, Yang, Xue, Zhang, Bai, and Qi}]{ding2023pla}
Ding, R.; Yang, J.; Xue, C.; Zhang, W.; Bai, S.; and Qi, X. 2023.
\newblock Pla: Language-driven open-vocabulary 3d scene understanding.
\newblock In \emph{Proceedings of the IEEE/CVF Conference on Computer Vision and Pattern Recognition}, 7010--7019.

\bibitem[{Feng et~al.(2021)Feng, Li, Li, Zhang, Zhang, Zhu, Zhang, Wang, and Mian}]{ffl-3dog}
Feng, M.; Li, Z.; Li, Q.; Zhang, L.; Zhang, X.; Zhu, G.; Zhang, H.; Wang, Y.; and Mian, A. 2021.
\newblock Free-form description guided 3d visual graph network for object grounding in point cloud.
\newblock In \emph{Proceedings of the IEEE/CVF International Conference on Computer Vision}, 3722--3731.

\bibitem[{Ge et~al.(2024)Ge, Yu, Zhao, Guo, Huang, Ren, Itti, and Wu}]{3dcopypaste}
Ge, Y.; Yu, H.-X.; Zhao, C.; Guo, Y.; Huang, X.; Ren, L.; Itti, L.; and Wu, J. 2024.
\newblock 3D Copy-Paste: Physically Plausible Object Insertion for Monocular 3D Detection.
\newblock \emph{Advances in Neural Information Processing Systems}, 36.

\bibitem[{Guo et~al.(2023)Guo, Tang, Zhang, Wang, Wang, Zhao, and Li}]{viewrefer}
Guo, Z.; Tang, Y.; Zhang, R.; Wang, D.; Wang, Z.; Zhao, B.; and Li, X. 2023.
\newblock Viewrefer: Grasp the multi-view knowledge for 3d visual grounding.
\newblock In \emph{Proceedings of the IEEE/CVF International Conference on Computer Vision}.

\bibitem[{Han et~al.(2024)Han, Zhao, Chen, Ma, and Zhang}]{han2024dual}
Han, Y.; Zhao, N.; Chen, W.; Ma, K.~T.; and Zhang, H. 2024.
\newblock Dual-Perspective Knowledge Enrichment for Semi-supervised 3D Object Detection.
\newblock In \emph{Proceedings of the AAAI Conference on Artificial Intelligence}.

\bibitem[{Hong et~al.(2023)Hong, Zhen, Chen, Zheng, Du, Chen, and Gan}]{3d-llm}
Hong, Y.; Zhen, H.; Chen, P.; Zheng, S.; Du, Y.; Chen, Z.; and Gan, C. 2023.
\newblock 3d-llm: Injecting the 3d world into large language models.
\newblock \emph{Advances in Neural Information Processing Systems}, 36: 20482--20494.

\bibitem[{Huang et~al.(2021)Huang, Lee, Chen, and Liu}]{tgNn}
Huang, P.-H.; Lee, H.-H.; Chen, H.-T.; and Liu, T.-L. 2021.
\newblock Text-guided graph neural networks for referring 3d instance segmentation.
\newblock In \emph{Proceedings of the AAAI Conference on Artificial Intelligence}, volume~35, 1610--1618.

\bibitem[{Huang et~al.(2022)Huang, Chen, Jia, and Wang}]{multiview-transformer}
Huang, S.; Chen, Y.; Jia, J.; and Wang, L. 2022.
\newblock Multi-view transformer for 3d visual grounding.
\newblock In \emph{Proceedings of the IEEE/CVF Conference on Computer Vision and Pattern Recognition}, 15524--15533.

\bibitem[{Jain et~al.(2022)Jain, Gkanatsios, Mediratta, and Fragkiadaki}]{butd-detr}
Jain, A.; Gkanatsios, N.; Mediratta, I.; and Fragkiadaki, K. 2022.
\newblock Bottom up top down detection transformers for language grounding in images and point clouds.
\newblock In \emph{European Conference on Computer Vision}, 417--433. Springer.

\bibitem[{Jiao et~al.(2024)Jiao, Zhao, Chen, and Jiang}]{jiao2024unlocking}
Jiao, P.; Zhao, N.; Chen, J.; and Jiang, Y.-G. 2024.
\newblock Unlocking textual and visual wisdom: Open-vocabulary 3d object detection enhanced by comprehensive guidance from text and image.
\newblock In \emph{European Conference on Computer Vision}, 376--392. Springer.

\bibitem[{Li et~al.(2023)Li, Li, Savarese, and Hoi}]{blip2}
Li, J.; Li, D.; Savarese, S.; and Hoi, S. 2023.
\newblock Blip-2: Bootstrapping language-image pre-training with frozen image encoders and large language models.
\newblock In \emph{International conference on machine learning}, 19730--19742. PMLR.

\bibitem[{Liu et~al.(2019)Liu, Ott, Goyal, Du, Joshi, Chen, Levy, Lewis, Zettlemoyer, and Stoyanov}]{liu2019roberta}
Liu, Y.; Ott, M.; Goyal, N.; Du, J.; Joshi, M.; Chen, D.; Levy, O.; Lewis, M.; Zettlemoyer, L.; and Stoyanov, V. 2019.
\newblock Roberta: A robustly optimized bert pretraining approach.
\newblock \emph{arXiv preprint arXiv:1907.11692}.

\bibitem[{Liu et~al.(2021)Liu, Zhang, Cao, Hu, and Tong}]{groupfree}
Liu, Z.; Zhang, Z.; Cao, Y.; Hu, H.; and Tong, X. 2021.
\newblock Group-free 3d object detection via transformers.
\newblock In \emph{Proceedings of the IEEE/CVF International Conference on Computer Vision}, 2949--2958.

\bibitem[{Luo et~al.(2025)Luo, Di, Yang, Ma, Xue, Wei, and Liu}]{Luo2025TraME}
Luo, C.; Di, D.; Yang, X.; Ma, Y.; Xue, Z.; Wei, C.; and Liu, Y. 2025.
\newblock TrAME: Trajectory-Anchored Multi-View Editing for Text-Guided 3D Gaussian Splatting Manipulation.
\newblock \emph{IEEE Transactions on Multimedia}.

\bibitem[{Luo et~al.(2022)Luo, Fu, Kong, Gao, Ren, Shen, Xia, and Liu}]{3d-sps}
Luo, J.; Fu, J.; Kong, X.; Gao, C.; Ren, H.; Shen, H.; Xia, H.; and Liu, S. 2022.
\newblock 3d-sps: Single-stage 3d visual grounding via referred point progressive selection.
\newblock In \emph{Proceedings of the IEEE/CVF Conference on Computer Vision and Pattern Recognition}, 16454--16463.

\bibitem[{Nekrasov et~al.(2021)Nekrasov, Schult, Litany, Leibe, and Engelmann}]{mix3d}
Nekrasov, A.; Schult, J.; Litany, O.; Leibe, B.; and Engelmann, F. 2021.
\newblock Mix3d: Out-of-context data augmentation for 3d scenes.
\newblock In \emph{2021 international conference on 3d vision (3dv)}, 116--125. IEEE.

\bibitem[{Pan et~al.(2024)Pan, Cao, Wang, Yang, and Wang}]{pan2024finding}
Pan, H.; Cao, Y.; Wang, X.; Yang, X.; and Wang, M. 2024.
\newblock Finding and Editing Multi-Modal Neurons in Pre-Trained Transformers.
\newblock In \emph{Findings of the Association for Computational Linguistics ACL 2024}, 1012--1037.

\bibitem[{Qi et~al.(2017)Qi, Yi, Su, and Guibas}]{pointnet++}
Qi, C.~R.; Yi, L.; Su, H.; and Guibas, L.~J. 2017.
\newblock Pointnet++: Deep hierarchical feature learning on point sets in a metric space.
\newblock \emph{Advances in neural information processing systems}, 30.

\bibitem[{Radford et~al.(2021)Radford, Kim, Hallacy, Ramesh, Goh, Agarwal, Sastry, Askell, Mishkin, Clark et~al.}]{clip}
Radford, A.; Kim, J.~W.; Hallacy, C.; Ramesh, A.; Goh, G.; Agarwal, S.; Sastry, G.; Askell, A.; Mishkin, P.; Clark, J.; et~al. 2021.
\newblock Learning transferable visual models from natural language supervision.
\newblock In \emph{International conference on machine learning}, 8748--8763. PMLR.

\bibitem[{Roh et~al.(2022)Roh, Desingh, Farhadi, and Fox}]{languagerefer}
Roh, J.; Desingh, K.; Farhadi, A.; and Fox, D. 2022.
\newblock Languagerefer: Spatial-language model for 3d visual grounding.
\newblock In \emph{Conference on Robot Learning}, 1046--1056. PMLR.

\bibitem[{Sheng et~al.(2022)Sheng, Cai, Zhao, Deng, Huang, Hua, Zhao, and Lee}]{sheng2022rethinking}
Sheng, H.; Cai, S.; Zhao, N.; Deng, B.; Huang, J.; Hua, X.-S.; Zhao, M.-J.; and Lee, G.~H. 2022.
\newblock Rethinking IoU-based optimization for single-stage 3D object detection.
\newblock In \emph{European Conference on Computer Vision}, 544--561. Springer.

\bibitem[{Vaswani et~al.(2017)Vaswani, Shazeer, Parmar, Uszkoreit, Jones, Gomez, Kaiser, and Polosukhin}]{transformer}
Vaswani, A.; Shazeer, N.; Parmar, N.; Uszkoreit, J.; Jones, L.; Gomez, A.~N.; Kaiser, {\L}.; and Polosukhin, I. 2017.
\newblock Attention is all you need.
\newblock \emph{Advances in neural information processing systems}, 30.

\bibitem[{Wang et~al.(2023)Wang, Huang, Zhao, Li, Cheng, Zhu, Yin, and Zhao}]{distill-weakly-3dvg}
Wang, Z.; Huang, H.; Zhao, Y.; Li, L.; Cheng, X.; Zhu, Y.; Yin, A.; and Zhao, Z. 2023.
\newblock Distilling coarse-to-fine semantic matching knowledge for weakly supervised 3d visual grounding.
\newblock In \emph{Proceedings of the IEEE/CVF International Conference on Computer Vision}, 2662--2671.

\bibitem[{Wu et~al.(2023)Wu, Cheng, Zhang, Cheng, and Zhang}]{eda}
Wu, Y.; Cheng, X.; Zhang, R.; Cheng, Z.; and Zhang, J. 2023.
\newblock Eda: Explicit text-decoupling and dense alignment for 3d visual grounding.
\newblock In \emph{Proceedings of the IEEE/CVF Conference on Computer Vision and Pattern Recognition}, 19231--19242.

\bibitem[{Yang et~al.(2024{\natexlab{a}})Yang, Zhang, Qi, Xu, Liu, Shan, Li, Yang, Li, Wang et~al.}]{core-3dvg}
Yang, L.; Zhang, Z.; Qi, Z.; Xu, Y.; Liu, W.; Shan, Y.; Li, B.; Yang, W.; Li, P.; Wang, Y.; et~al. 2024{\natexlab{a}}.
\newblock Exploiting Contextual Objects and Relations for 3D Visual Grounding.
\newblock \emph{Advances in Neural Information Processing Systems}, 36.

\bibitem[{Yang et~al.(2021{\natexlab{a}})Yang, Feng, Ji, Wang, and Chua}]{yang2021deconfounded}
Yang, X.; Feng, F.; Ji, W.; Wang, M.; and Chua, T.-S. 2021{\natexlab{a}}.
\newblock Deconfounded video moment retrieval with causal intervention.
\newblock In \emph{Proceedings of the 44th International ACM SIGIR Conference on Research and Development in Information Retrieval}, 1--10.

\bibitem[{Yang et~al.(2022)Yang, Wang, Dong, Dong, Wang, and Chua}]{yang2022video}
Yang, X.; Wang, S.; Dong, J.; Dong, J.; Wang, M.; and Chua, T.-S. 2022.
\newblock Video moment retrieval with cross-modal neural architecture search.
\newblock \emph{IEEE Transactions on Image Processing}, 31: 1204--1216.

\bibitem[{Yang et~al.(2024{\natexlab{b}})Yang, Zeng, Guo, Wang, Dong, and Wang}]{yang2024robust}
Yang, X.; Zeng, J.; Guo, D.; Wang, S.; Dong, J.; and Wang, M. 2024{\natexlab{b}}.
\newblock Robust Video Question Answering via Contrastive Cross-Modality Representation Learning.
\newblock \emph{SCIENCE CHINA Information Sciences}, 67: 1--16.

\bibitem[{Yang et~al.(2021{\natexlab{b}})Yang, Zhang, Wang, and Luo}]{sat}
Yang, Z.; Zhang, S.; Wang, L.; and Luo, J. 2021{\natexlab{b}}.
\newblock Sat: 2d semantics assisted training for 3d visual grounding.
\newblock In \emph{Proceedings of the IEEE/CVF International Conference on Computer Vision}, 1856--1866.

\bibitem[{Yuan et~al.(2021)Yuan, Yan, Liao, Zhang, Wang, Li, and Cui}]{instancerefer}
Yuan, Z.; Yan, X.; Liao, Y.; Zhang, R.; Wang, S.; Li, Z.; and Cui, S. 2021.
\newblock Instancerefer: Cooperative holistic understanding for visual grounding on point clouds through instance multi-level contextual referring.
\newblock In \emph{Proceedings of the IEEE/CVF International Conference on Computer Vision}, 1791--1800.

\bibitem[{Zhang, Wang, and Loy(2020)}]{outdooraugmentation}
Zhang, W.; Wang, Z.; and Loy, C.~C. 2020.
\newblock Exploring data augmentation for multi-modality 3d object detection.
\newblock \emph{arXiv preprint arXiv:2012.12741}.

\bibitem[{Zhang, Gong, and Chang(2023)}]{multi3drefer}
Zhang, Y.; Gong, Z.; and Chang, A.~X. 2023.
\newblock Multi3drefer: Grounding text description to multiple 3d objects.
\newblock In \emph{Proceedings of the IEEE/CVF International Conference on Computer Vision}, 15225--15236.

\bibitem[{Zhao et~al.(2021)Zhao, Cai, Sheng, and Xu}]{3dvg-transformer}
Zhao, L.; Cai, D.; Sheng, L.; and Xu, D. 2021.
\newblock 3DVG-Transformer: Relation modeling for visual grounding on point clouds.
\newblock In \emph{Proceedings of the IEEE/CVF International Conference on Computer Vision}, 2928--2937.

\bibitem[{Zhao, Chua, and Lee(2020)}]{zhao2020sess}
Zhao, N.; Chua, T.-S.; and Lee, G.~H. 2020.
\newblock Sess: Self-ensembling semi-supervised 3d object detection.
\newblock In \emph{Proceedings of the IEEE/CVF Conference on Computer Vision and Pattern Recognition}, 11079--11087.

\bibitem[{Zhao and Lee(2022)}]{zhao2022static}
Zhao, N.; and Lee, G.~H. 2022.
\newblock Static-dynamic co-teaching for class-incremental 3d object detection.
\newblock In \emph{Proceedings of the AAAI Conference on Artificial Intelligence}, volume~36, 3436--3445.

\bibitem[{Zhao, Zhao, and Lee(2022)}]{zhao2022synthetic}
Zhao, Y.; Zhao, N.; and Lee, G.~H. 2022.
\newblock Synthetic-to-Real Domain Generalized Semantic Segmentation for 3D Indoor Point Clouds.
\newblock \emph{arXiv preprint arXiv:2212.04668}.

\end{thebibliography}
